\documentclass[letterpaper, 10pt, conference]{IEEEtran}      

\IEEEoverridecommandlockouts                              

\IEEEpubid{\begin{minipage}{\columnwidth}\copyright 2022 IEEE.  Personal use of this material is permitted. Permission from IEEE must be obtained for all other uses, in any current or future media, including reprinting/republishing this material for advertising or promotional purposes, creating new collective works, for resale or redistribution to servers or lists, or reuse of any copyrighted component of this work in other works.
\end{minipage}
\hspace{\columnsep}\makebox[\columnwidth]{ }
}

\usepackage{amsmath}
\usepackage{subfigure}
\usepackage{lipsum}
\usepackage[]{graphicx}
\usepackage[dvipsnames]{xcolor}
\usepackage{booktabs, cellspace}
\usepackage{algorithm}
\usepackage{algpseudocode}
 \usepackage[normalem]{ulem}
 
\title{\LARGE \bf Affordance detection with Dynamic-Tree Capsule Networks}

\author{A. Rodr\'iguez-S\'anchez$^{1}$ \and S. Haller-Seeber$^{1}$ \and D. Peer$^{1}$ \and C. Engelhardt$^{1}$ \and J. Mittelberger$^{1}$ \and M. Saveriano$^{2}$ 
\thanks{$^{1}$Department of Computer Science, University of Innsbruck, Innsbruck, Austria {\tt antonio.rodriguez-sanchez@uibk.ac.at}}%
\thanks{$^{2}$Department of Industrial Engineering, University of Trento, Italy.}%
}

\begin{document}

\maketitle
\IEEEpubidadjcol

\begin{abstract}
Affordance detection from visual input is a fundamental step in autonomous robotic manipulation. Existing solutions to the problem of affordance detection rely on convolutional neural networks. However, these networks do not consider the spatial arrangement of the input data and miss parts-to-whole relationships. Therefore, they fall short when confronted with novel, previously unseen object instances or new viewpoints. One solution to overcome such limitations can be to resort to capsule networks. In this paper, we introduce the first affordance detection network based on dynamic tree-structured capsules for sparse 3D point clouds.
We show that our capsule-based network outperforms current state-of-the-art models on viewpoint invariance and parts-segmentation of new object instances through a novel dataset we only used for evaluation and it is publicly available from  \texttt{github.com/gipfelen/DTCG-Net}. In the experimental evaluation we will show that our algorithm is superior to current affordance detection methods when faced with grasping previously unseen objects thanks to our Capsule Network enforcing a parts-to-whole representation.
\end{abstract}

%
%
%
\section{\label{chap:introduction}Introduction}

3D sensing is essential in robotics and is one of the major research areas in the field. Thanks to the recent advances of deep learning, much progress has been achieved in many challenging tasks like object classification or affordance detection ~\cite{do2018affordancenet,DCG-Net}. Existing approaches typically rely on convolutional neural networks (CNN) due to their outstanding performance in most vision applications. However, CNN-based networks consider the spatial arrangement the input data only based on local proximity and not in terms of the arrangement of constituent parts as they do not explicitly encode parts-to-whole relationships. The consequence of this limitation is a drop in performance when faced with new object instances or new viewpoints.

An evolution of CNNs are Capsule Networks that group multiple neurons into a single Capsule which are layer-wise connected through a special routing-procedure~\cite{DBLP:journals/corr/abs-1710-09829}. Recently, Capsule Networks have been successfully applied to 3D data~\cite{zhao20193d} for tasks like reconstruction, segmentation, and interpolation. 
Goyal et al.~\cite{indcutive_bias_dl} already motivated that  additional inductive biases should be added to deep learning methods in order to go from a reasonably good \textit{in-distribution generalization} in highly supervised learning tasks to strong \textit{out-of-distribution generalization}. This has been additionally supported by recent work ~\cite{dynamicTreeCapsulesPointCloud}, which introduced the auto-encocder DT-CapsNet, outperforming routing-by-agreement (RBA) in terms of reconstruction quality and classification accuracy. In this paper, we exploit a special routing procedure and architecture that implements an inductive bias that enforces a parts-to-whole object relationship between different layers to improve out-of-distribution generalization. It will be demonstrated experimentally that such inductive bias helps to correctly detect objects that were excluded from training and greatly improves scores when compared to the state-of-the-art CNNs. When it comes to affordances~\cite{min2016affordance, zech2017computational}, which in robotics are typically associated with object parts (e.g., the handle of a cup affords for \textit{grasp}), the introduced inductive bias significantly improves the detection even on unforeseen objects. We show experimentally that such better out-of-distribution generalization results in higher affordances success rates when determining the right grasping pose for an object.  


The contribution of this paper is threefold. \textit{i)} we introduce the first capsule-based affordance detection network which enforces a part-to-whole object relationship into capsules. 
\textit{ii)} we experimentally show that Capsule Networks provide comparable results to CNNs when the training and test data are similar, but more importantly, we show that the proposed approach outperforms CNNs in terms of out-of-distribution generalization. \textit{iii)} in order to show such superiority, we introduce a newly created dataset (namely UIBK dataset) which augments the UMD dataset \cite{Myers:ICRA15} by adding $40$ RGB-D images of $7$ different kitchen tools. Neither the new tool instances, nor the novel viewpoints are included in the original UMD dataset. The UIBK dataset is publicly available at \texttt{github.com/gipfelen/DTCG-Net}.

After presenting the related work in section \ref{chap:related_work} reviews, we introduce in section \ref{chap:method} our Dynamic-Tree Capsule Graph Network (DTCG-Net), which builds on DCG-Net \cite{DCG-Net} and Scale Distance Routing (SDA) \cite{peer2018training}. In section \ref{chap:experiments}, we compare our DTCG-Net with a Capsule Network using Routing by Agreement (RBA)~\cite{DBLP:journals/corr/abs-1710-09829} and a state-of-the-art CNN network~\cite{wang2019dynamic}. Section \ref{chap:robot} presents an experiment where affordance detection is used to let a real robot grasp previously unseen kitchen tools, providing the success rates of our approach and the aforementioned competitor networks. Finally, in section \ref{chap:conclusion}, we state the conclusions and provide insights about possible improvements and further research directions. 

\section{\label{chap:related_work}Related Work}

\begin{figure*}
    \begin{center}
    \includegraphics[width=\linewidth]{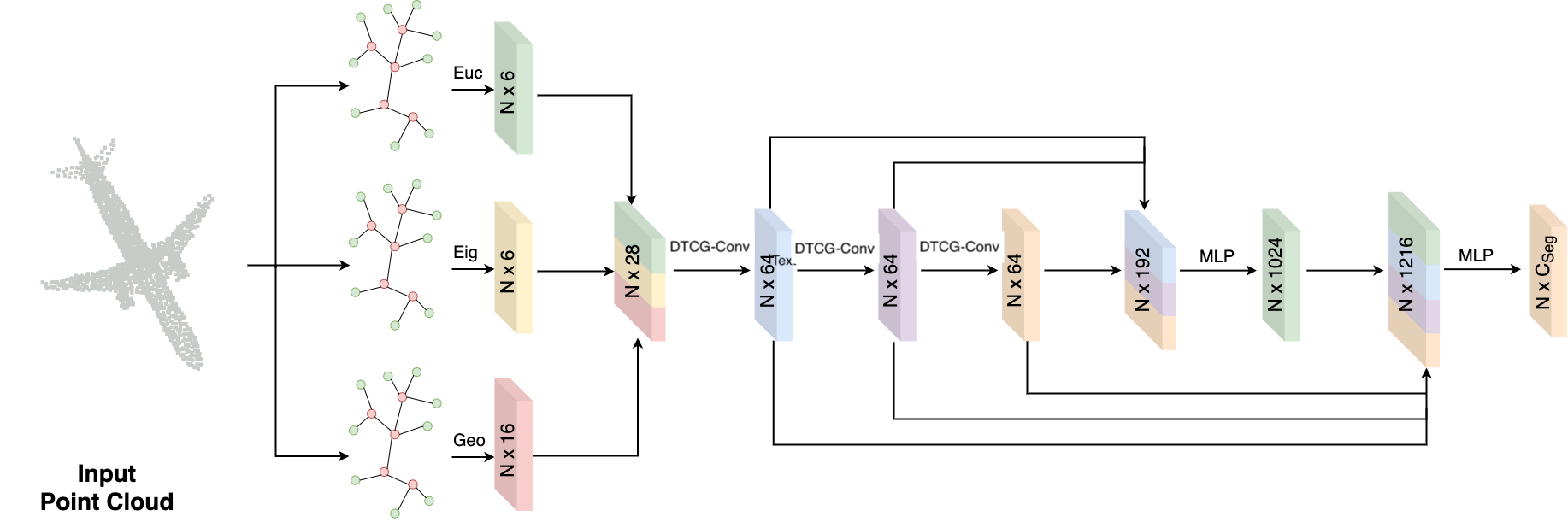}
    \end{center}
       \caption{Visualization of the DTCG-Net. It first extracts the eigenvalues and euclidean and geometrical features from the input point cloud. These will pass through multiple DTCG-Conv~\cite{DCG-Net} modules for the prediction of the part segmentation.}
    \label{fig:architecture}
\end{figure*}
%
Capsule Networks~\cite{DBLP:journals/corr/abs-1710-09829} are an evolution of Convolutional Neural Networks (CNNs), where the main component is the \textit{capsule}. In this context, and differently from neurons in CNNs, a capsule is a vector, representing a part of an object or an entity from a class of objects. Even though Capsule Networks are trained by backpropagation, the fact that capsules are represented by vectors, information is routed by an iterative algorithm. The first routing algorithm was Routing by Agreement (RBA)~\cite{DBLP:journals/corr/abs-1710-09829} that introduced a coupling coefficient that would represent probabilities regarding the agreement of lower layer capsules with capsules in the next capsule layer. Later, Hinton and colleagues~\cite{e2018matrix} applied Expectation-Maximization in order to compute this routing. Recent routing algorithms have applied other methodologies, such as distance scaling\cite{peer2019gammacapsules}, attention~\cite{tsai2020capsules} or the use of variational bayes~\cite{ribeiro2020capsule}.

Capsule Networks have been widely applied, applications range from detecting actions in videos~\cite{DBLP:journals/corr/abs-1805-08162} to multitask learning~\cite{xiao2018mcapsnet}, medical applications~\cite{zhang2021onfocus,afshar2020covid}, or 3D point clouds, the subject of this work. On the latter, one of the first works involved an auto-encoder aimed at learning 3D unstructured data ~\cite{zhao20193d} that showed its success in parts segmentation, object reconstruction and classification. The more recent DCG-Net~\cite{DCG-Net} has been applied to these same  applications, being comparable to non-capsule-based approaches such as DG-CNN~\cite{wang2019dynamic}, GS-Net~\cite{xu2020geometry}, PointNet~\cite{qi2017pointnet}, or PointNet++~\cite{qi2017pointnet++}. In regards to other applications of Capsule Networks in 3D point clouds, these include hand-pose estimation\cite{sym12101636} or edge detection~\cite{app11041833}. In relation to such tasks is the affordance estimation of parts of objects\cite{do2018affordancenet,chu2019recognizing}, where Convolutional Neural Networks~\cite{qi2017pointnet++,wang2019dynamic,deng20213d} can be applied to segment parts of objects that correspond to the same functionality and would then have similar affordances. We will show next that since capsules aim at representing parts of objects, it is only natural to apply Capsule Networks to such tasks.



%
%
%

\section{\label{chap:method}Methods}

We present here the details of our inductive bias (Sec.~\ref{chap:introduction}) for Capsule Networks on 3D point clouds to improve out-of-distribution generalization. More precisely, with our approach, we enforce such inductive bias by generating a tree-like structure during inference. Including an inductive bias allows for  \textit{out-of-distribution generalization}~\cite{indcutive_bias_dl, dynamicTreeCapsulesPointCloud}. Affordances are very much related to object parts, as an example, handles are for \textit{grasp} affordance. The inductive routing algorithm we apply~\cite{peer2019gammacapsules} (section \ref{chap:routing}) enforces a tree-like structure, which is not always the case when using other routing algorithms such as RBA or EM. 

In our Capsule Network, shown in Fig.~\ref{fig:architecture} and named \emph{Dynamic-Tree Capsule Network} (DTCG-Net), each capsule embodies part of an object or an object instance. We will experimentally show that enforcing this parts-to-whole object relationship improves the generalization capability of Capsule Networks.  


\subsection{Network Architecture}
\label{chap:architecture}
DTCG-Net's architecture follows on DCG-Net~\cite{DCG-Net}, updated for a dynamic tree representation. The network uses multiple Dynamic-Tree Capsule Graph Convolutions (DTCG-Conv) modules as follows. The first DTCG-Conv layer is fed with the 3D point cloud extracting features from an eigenvalue,  geometrical and euclidean nature, which are concatenated, giving rise to 28 channels as shown in Fig.~\ref{fig:architecture}. Such concatenated features are the input to the next DTCG-Conv layer, to which follows two additional DTCG-Conv layers. Each of these three DTCG-Conv layers are batch normalized ~\cite{DBLP:journals/corr/IoffeS15} and non-linearized using leaky-ReLU ~\cite{DBLP:journals/corr/XuWCL15}, the slope for negative values is set to 0.2. Next, as shown in Fig.~\ref{fig:architecture}, through the use of skip-connections, the outputs from these three DTCG-Conv layers are accumulated into a layer that will serve as input to a fully-connected Multi-layer Perceptron (MLP) which finally provides the class label of each point that belongs to the point cloud.


\subsection{\label{chap:routing}Routing between capsules}
In our DTCG-Net, we enforce capsules to represent a whole object or parts of it. This is of interest for affordance detection from point cloud data as capsules would represent point clouds relevant for parts representation, as opposed to methods such as CNN where the focus is on salient features leading to high-classification rates. To successfully implement this inductive bias, each capsule of the network should represent an object instance or one of its component parts during inference. The most usual routing algorithms do not always represent a part's whole relationship~\cite{peer2019gammacapsules}. If we are to use vector distances instead of the vector product present in routing-by-agreement (RBA), capsules represent the object part in a dynamic tree to a much higher degree as shown in 2D images~\cite{peer2019gammacapsules}. We followed this idea of using vector distances in the routing algorithm in order to enforce a parts-to-whole relationships for 3D point clouds and Capsule Graph Convolutional layers. The pseudo-code is shown in Algorithm~\ref{algo:sda}. Line 3 of Algorithm~\ref{algo:sda} restricts the prediction vectors to the activation of its predicting capsule. This ensures that a lower level capsule can only activate an upper level capsule when the correlated feature of this capsule is present in the current input. Line 4 represents the calculation of the coupling coefficient, which is used to calculate the output of higher level capsules in lines 5 and 6. To ensure a strong parse tree structure with a large coupling coefficient the distance is multiplied with the scale factor $t$ to get the scaled distance agreement (line 9).

\begin{algorithm}[t]
   \caption{Scale-distance-agreement (SDA) ~\cite{peer2019gammacapsules} is the algorithm used for DTCG-Net Routing: Computes the coupling $c_{ij}$ between the $i$ capsules in layer $l$ 
   with the $j$ capsules in layer $l+1$ imposing our inductive bias (see text). Input are the predictions $\hat{u}_{j|i}$ from lower level capsule activations $v_i$, the number of iterations $r$ and the layer $l$}\label{algo:sda} 
   \begin{algorithmic}[1]
     \Procedure{SdaRouting}{$v_i$,$\hat{u}_{j|i},r,l$}  
     \State $b_{ij} \gets 0$
     \State $\hat{u}_{j|i} \gets \min(||v_i||, ||\hat{u}_{j|i}||) \frac{\hat{u}_{j|i}}{||\hat{u}_{j|i}||} $ \label{algo:sda_line:restrict}
     \For{$r \text{ iterations}$}
       \State $c_{ij} \gets \frac{\exp(b_{ij})}{\sum_k \exp(b_{ik})}$ \label{algo:sda:line:cij}
       \State $s_j \gets \sum_{i} c_{ij}\hat{u}_{j|i}$ \label{algo:sda:line:sj} 
       \State $v_j \gets \frac{||s_j||^2}{1+||s_j||^2} \frac{s_j}{||s_j||}$ \label{algo:sda:line:vj} 
       \State $t_i \gets \frac {\log(0.9 (J-1)) - \log(1 - 0.9)} {-0.5 \mathrm{mean}^J_j(||\hat{u}_{j|i} - v_j||)}$ \label{algo:sda:line:t} 
       \State $b_{ij} \gets ||\hat{u}_{j|i} - v_j|| t_i$\label{algo:sda:line:bij} 
     \EndFor
     \EndProcedure
   \end{algorithmic}
 \end{algorithm}





\subsection{\label{chap:metrics}Evaluation Metrics}
To compare the different networks we used 2 different scores, namely the accuracy and the  mean  intersection  over  union  (mIoU).

Our network gets as input a point cloud $\mathcal{P} = \{p_i\}_{i=1}^N$ with $N$ points and assigns to each point a class label $l_i$ (see Sec.~\ref{chap:architecture}). We compare the network's output labels ($l_i$) with the ground truth labels ($g_i$) to calculate the accuracy as:
\begin{equation}
\label{math:accuracy}
accuracy = \frac{\sum_{i=1}^N \delta(l_i,g_i)}{N},
\end{equation}
where $\delta(x,y)$ is the  Kronecker delta function. If $x\ne y$, then $\delta(x,y)=0$ and $\delta(x,y)=1$ if $x=y$.
By definition, the accuracy gives us a measure of how many points have been labeled correctly and thus how good the segmentation is.

An additional metric to measure the segmentation quality is the mean intersection over union (mIoU)~\cite{wang2019dynamic}. It is the average of all the intersections over union (IoU). These are calculated by dividing the number of correctly classified points per segmentation category ($c$) by the number of points, where either the network prediction or the ground truth label has this segmentation category:
\begin{equation}
IoU_{c} = \frac{\sum_{i=1}^N \delta(g_i,c) \cdot \delta(l_i,c)}{\sum_{i=1}^N \delta(g_i,c) + \delta(l_i,c)}. 
\label{math:IoU}
\end{equation}

\section{\label{chap:experiments}Experimental Evaluation}
\subsection{\label{chap:dataset}Datasets}
    We perform our experimental evaluation on three datasets:
    \begin{itemize}
        \item The UMD Part Affordance dataset~\cite{Myers:ICRA15} contains RGB-D images for 105 gardening, kitchen and workshop tools from different viewpoints which are segmented into 7 different affordances. We used this dataset for training and evaluation.
        \item The 3D AffordanceNet (AffNet) dataset~\cite{deng20213d} is composed of 23000 shapes corresponding to 23 objects. In terms of affordances, such objects were annotated with 18 visual affordance categories. Annotations are of a multi-class and multi-label nature, meaning that each individual point can have multiple affordance labels. We used the affordance class with the largest value for each point so as to be in line with the previous dataset which provides only one label per point.
        \item Our UIBK dataset 
        contains
        40 RGB-D images of 7 different kitchen tools (\texttt{scissor}, \texttt{mug}, \texttt{knife}, \texttt{spoon}, \texttt{turner}, \texttt{hammer}, \texttt{bowl}), captured with an Intel Realsense Camera D415 and converted to RGB-D images~\cite{librealsense}. 
        This dataset contains additional unseen viewpoints like a top view and was used to evaluate unseen object instances as well as novel viewpoints. We increased the size of the dataset adding 360 images obtained by rotating and scaling the original ones. We repeated the augmentation twice, obtaining the UIBK-20 (rotation between $\pm 10\,$deg and scaling between $\pm 10\,$cm) and the UIBK-40 (rotation between $\pm 20\,$deg and scaling between $\pm 20\,$cm) dataset. 
        We used the UIBK, UIBK-20 and UIBK-40 datasets only for evaluation, that is, we did not use this dataset for training in order to show the generalization capabilities of DTCG-Net.
    \end{itemize}

\subsection{\label{chap:setup}Setup}
%
%
\begin{table}[t]
    \centering
    \caption{Affordance (A) Color (C) mapping.}
    \label{tab:affordances}
    {\resizebox{\columnwidth}{!}{
 	\begin{tabular}{cccccccc}
 	\toprule
      \sc{A} & Grasp & Cut & Scoop & Contain & Pound & Support & Wrap Grasp \\\midrule
       \sc{C} & Blue & Orange & Green & Red & Violet & Brown & Pink \\\bottomrule
    \end{tabular}
    } }
\end{table}
RGB-D images in the considered datasets contain much background information that we remove using the ground truth labels. Since the tested neural networks require point clouds as input, we convert RGB-D images (without background) into point clouds with the library Open3D~\cite{Zhou2018}. 
For clarity in the evaluation, every affordance is attached a color as shown in Tab.~\ref{tab:affordances}. 
    
    We trained the Capsule based networks, both DTCG-Net and DCG-Net, for 200 epochs. We used an adaptive MultiStepLR learning rate~\cite{deng20213d} with an initial learning rate of $lr=0.0001$, the decay rate was set to $\gamma=0.1$ and two milestones were defined at $75$ and $125$. We included in the comparison the CNN based architecture DG-CNN. Even though there are other 3D point cloud CNN architectures ~\cite{thomas2019kpconv,chen2021equivariant}, we chose DG-CNN for its popularity, its performance, and more importantly, its ability to provide object parts, which is essential for our study. For DG-CNN, we used exponential decay learning rate~\cite{NEURIPS2019_9015} with an initial learning rate of $lr=0.003$, a decay step of $337.620$ (as defined in their git repository~\cite{dgcnngit}), decay rate was set to $\gamma=0.5$.
    The batch size for all networks was 16 and all networks were trained on an Nvidia RTX 3090.

\subsection{\label{chap:qualitative_analysis}Results}
\begin{figure}
    \centering
    \begin{tabular}{ccc}
        \footnotesize{\sc{DG-CNN}} & \footnotesize{\sc{DCG-Net}} & \footnotesize{\sc{DTCG-Net (Ours)}} \\
        \includegraphics[width = 0.9in]{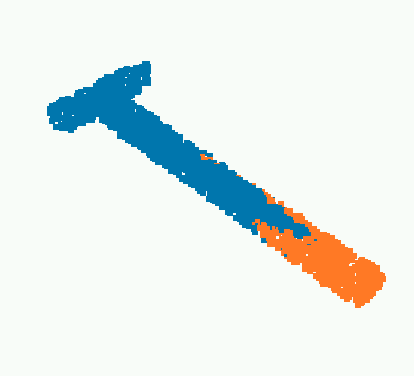} & \includegraphics[width = 0.9in]{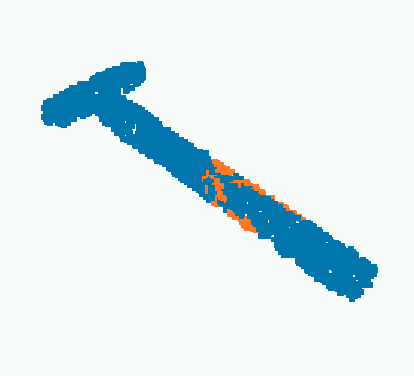} & \includegraphics[width = 0.9in]{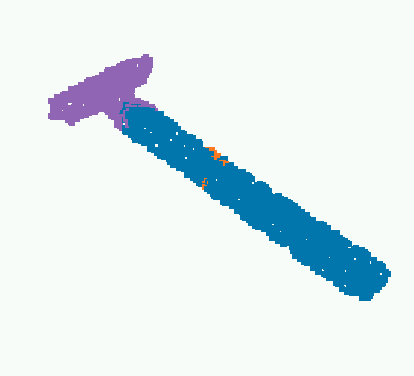} \\
        \footnotesize{$(50.93|68.87)$} & \footnotesize{$(73.63|71.70)$} & \footnotesize{$(93.95|83.87)$} \\
        & \footnotesize{(a)} & \\
        \includegraphics[width = 0.9in]{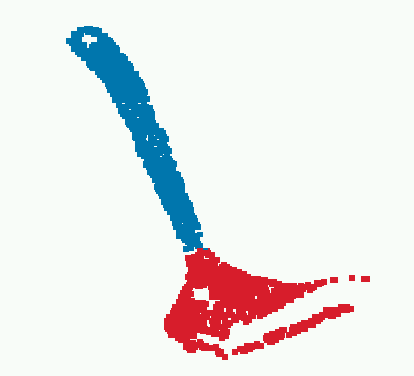} & \includegraphics[width = 0.9in]{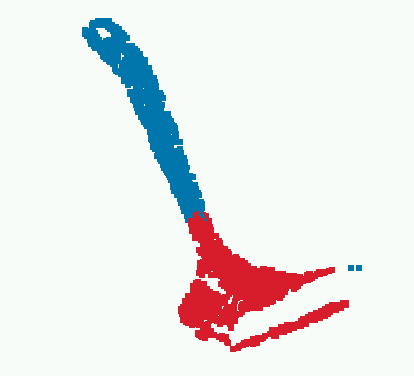} & \includegraphics[width = 0.9in]{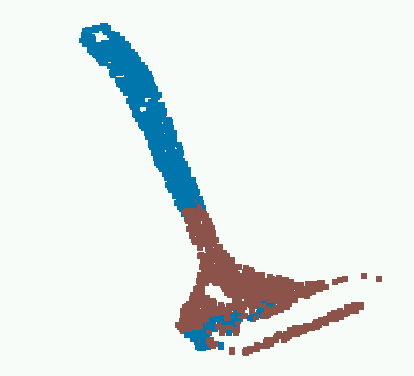} \\
        \footnotesize{$(50.44|74.99)$} & \footnotesize{$(46.24|73.93)$} & \footnotesize{$(88.81|94.97)$} \\
        & \footnotesize{(b)} & \\
        \includegraphics[width = 0.9in]{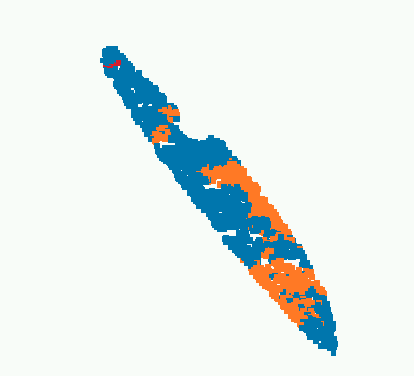} & \includegraphics[width = 0.9in]{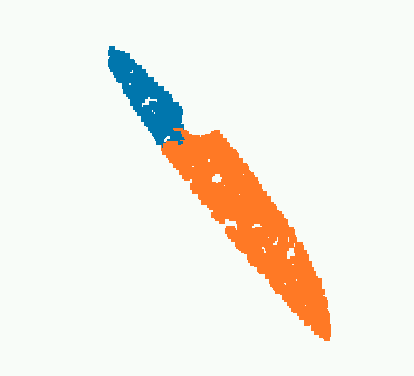} & \includegraphics[width = 0.9in]{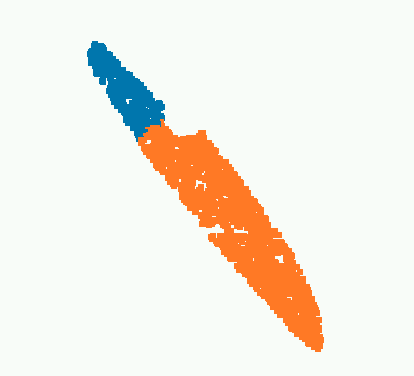} \\
        \footnotesize{$(48.19|70.34)$} & \footnotesize{$(98.59|98.98)$} & \footnotesize{$(96.14|97.23)$} \\
        & \footnotesize{(c)} & \\
        \includegraphics[width = 0.9in]{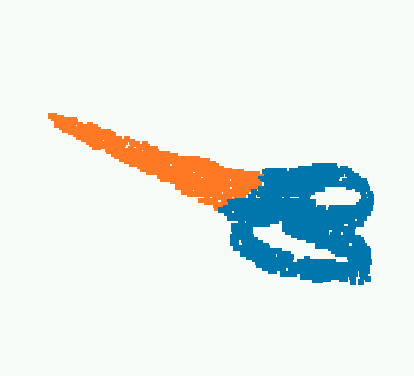} & \includegraphics[width = 0.9in]{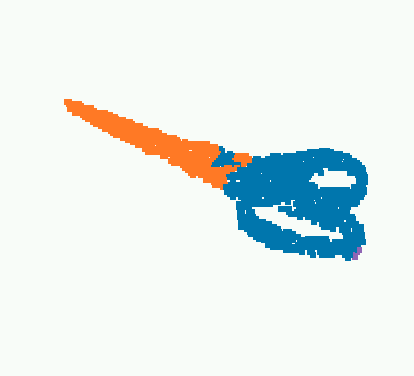} & \includegraphics[width = 0.9in]{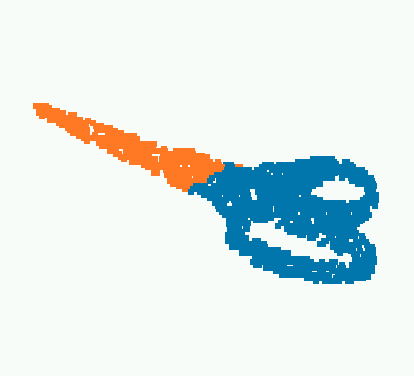} \\
        \footnotesize{$(84.57|91.22)$} & \footnotesize{$(87.40|80.55)$} & \footnotesize{$(90.14|93.89)$} \\
        & \footnotesize{(d)} & \\
        \includegraphics[width = 0.9in]{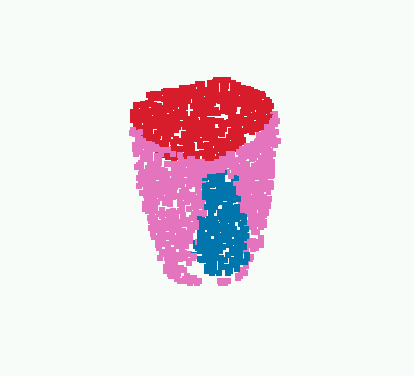} & \includegraphics[width = 0.9in]{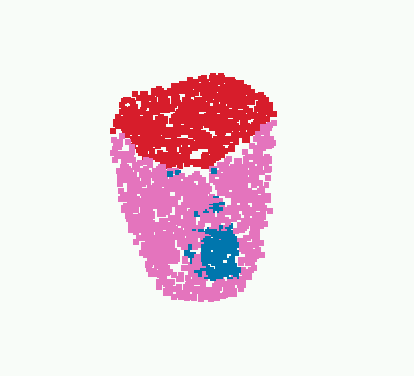} & \includegraphics[width = 0.9in]{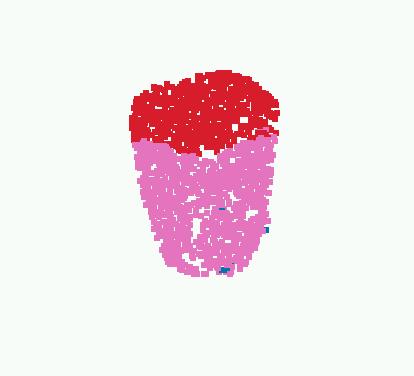} \\
        \footnotesize{$(92.77|93.41)$} & \footnotesize{$(89.31|89.41)$} & \footnotesize{$(86.72|84.39)$} \\
        & \footnotesize{(e)} & \\
    \end{tabular}
\caption{Segmentation output of the different networks with unseen tools from UIBK dataset: a) \texttt{hammer}, b) \texttt{turner}, c) \texttt{knife}, d) \texttt{scissors}, and e) \texttt{mug}. We report (accuracy [\%] $|$ mIoU [\%]) under each image.
\label{fig:compare}}
\end{figure}
\begin{table*}[t]
    \centering
    \caption{Accuracy and mIoU for the three different networks on the AffNet \cite{deng20213d}, the UMD \cite{Myers:ICRA15}, and our datasets.}
    \label{tab:accuracy:iou}
    {\renewcommand\arraystretch{1.2} 
 	\begin{tabular}{cccccccc}
 	\toprule
    \multicolumn{1}{c}{}& \multicolumn{1}{c}{}& \multicolumn{2}{c}{\sc{DG-CNN}}  & \multicolumn{2}{c}{\sc{DCG-Net}} & \multicolumn{2}{c}{\sc{DTCG-Net}} \\
        \sc{Test Dataset} & \sc{Train Dataset}    & \sc{Accuracy [\%]} & \sc{mIoU [\%]}     & \sc{Accuracy [\%]} & \sc{mIoU [\%]}  & \sc{Accuracy [\%]} & \sc{mIoU [\%]}       \\\midrule
        \sc{UMD}     & \sc{UMD}   & 97.46  & 98.09 &   95.24         &   96.30     &    95.17       &    96.18           \\
        \sc{AffNet}   & \sc{AffNet}  & 67.80  & 89.94 &   61.78         &   88.65     &    64.78        &    89.24           \\
        \sc{UIBK} & \sc{UMD}   &   72.12         &   84.52        &   75.48         &   84.41     &  76.93 &  85.84 \\    
        \sc{UIBK-20} & \sc{UMD} &   69.31         &   82.09        &   70.28         &   81.49     &  71.42 &  82.41    \\    
        \sc{UIBK-40} & \sc{UMD} &     59.42    &    75.98     &       67.81     &    73.34   &  65.59 &  78.71    \\\bottomrule \end{tabular}}
\end{table*}

Results are shown in Tab.~\ref{tab:accuracy:iou} as well as in Fig.~\ref{fig:compare}. Looking at UMD, we can see that the CNN-based network DG-CNN outperforms (in terms of both accuracy and mIoU) their capsule network counterparts by around 2\% (Tab.~\ref{tab:accuracy:iou}), while these last ones, DCG-Net as well as DTCG-Net have similar results. The same holds for AffNet, where DG-CNN has again the highest scores for the mIoU as well as for the accuracy. However, in this case DTCG-Net outperforms DCG-Net with a higher accuracy of around 3\%. The mIoU is nearly the same to all three networks, where DTCG-Net is very close to DG-CNN. Thus the CNN is still the go-to method to achieve the best performance when evaluating on samples from the same dataset used for training.

Figure~\ref{fig:compare} shows the affordance output for $5$ different unseen tools from UIBK dataset generated by the three different networks. We also report the accuracy and mIoU in percentages below each image. The average accuracy and mIoU over the entire dataset are listed in Tab.~\ref{tab:accuracy:iou}. It is very important to note that, while for the UMD and AffNet we divided the dataset into a training and test sets, our datasets (UIBK, UIBK-20 and UIBK-40) was only used for testing and thus was completely unseen to all the evaluated networks (DG-CNN, DCG-Net and DTCG-Net) in order to evaluate the out-of-distribution capabilities of each network. The weights from the UMD dataset were reused since it contains the exact same tools and affordances as described in Sec.~\ref{chap:dataset}. The main goal of our DTCG-Net is to learn more general object features, which would show in reaching higher scores in scenarios of newly unseen viewpoints or novel data. The left column corresponds to the DG-CNN network, the middle column to the capsule based DCG-Net with RBA as routing and the right column to our capsule based DTCG-Net with the SDA routing. 

Focusing on specific objects and affordances, for example the \texttt{hammer}, our network is the only one that labels the head as \texttt{pound} (violet points in Fig.~\ref{fig:compare}a). In this case, the DG-CNN network not only fails at detecting the pounding affordance, but also annotates a part of the handle with the \texttt{cut} (orange) label, achieving an accuracy of just 50.93\%. DCG-Net also fails to detect the pounding part, but it achieves a higher accuracy of 73.63\%. Our DTCG-Net is the most accurate (93.95\%). The same can be said about the \texttt{turner}, where DG-CNN and DCG-Net label the bottom part as \texttt{contain} (red), DTCG-Net labels it as \texttt{support} (brown). This results in low accuracies for DG-CNN and DCG-Net (accuracies of 50.44\% and 46.24\% respectively), while DTCG-Net did the segmentation properly with an accuracy of 88.81\%.  DTCG-Net also gets similar results to DCG-Net for the \texttt{knife} with both networks that reasonably label \texttt{grasp} (blue) and \texttt{cut} (orange) with an accuracy of 96.14\% and 98.58\% respectively. On the same object, DG-CNN performs poorly  (accuracy of 48.19\%) mixing up \texttt{grasp} and \texttt{cut} points. For the \texttt{scissors} (Fig.~\ref{fig:compare}d), all three networks performed similarly well (accuracy of 84.57\%, 87.40\% and 90.14\% for DG-CNN, DCG-Net and DTCG-Net respectively). This is probably due to the fact that we could incorporate less variability to this object and the views obtained were similar to  the ones present in the UMD dataset~\cite{Myers:ICRA15}.
DTCG-Net performs well on all the objects, apart from the \texttt{mug} (accuracy of 86.72\%) where it almost missed the handle (Fig.~\ref{fig:compare}e). On the same object, DG-CNN is the most accurate followed by DCG-Net (accuracy of 92.77\% and 89.31\% respectively). This behavior is probably due to the inductive bias enforced in DTCG-Net that tends to segment the object into fewer but larger segments. While this behavior provides a clear advantage for object segmentation---thanks to a more clear separation between segments---leading to better affordance labeling in most cases, it has the downside of neglecting parts that are too small. To summarize, 
Fig.~\ref{fig:compare} qualitatively shows several examples where our approach (DTCG-Net) clearly outperforms DCG-Net and DG-CNN (Fig.~\ref{fig:compare}a to~\ref{fig:compare}c), provides similar results (Fig.~\ref{fig:compare}d), and slightly worse values (Fig.~\ref{fig:compare}e). For a more rigorous analysis, we show in Tab.~\ref{tab:accuracy:iou} the accuracy and the mIoU  (Sec.~\ref{chap:metrics}) for UMD, AffNet, and UIBK (-20, -40) datasets evaluated on all three networks.  

Analyzing the results shown in Tab.~\ref{tab:accuracy:iou} we notice that, if we are to evaluate the three networks with previously not seen data (our three UIBK datasets), DG-CNN provides now the worst results, while the capsule-based networks hold stronger. For the UIBK dataset (40 images), DTCG-Net shows the best accuracy (76.93\%) and mIoU (85.84\%), compared with DCG-Net (accuracy of 75.48\% and mIoU 84.41\%) and DG-CNN (accuracy of 72.12\% and mIoU 84.51\%). While the accuracy of the DTCG-Net drops by around 18\%, the accuracy of DG-CNN drops by around 25\%. The same holds true for the mIoU where the DG-CNN drops by around 4\% more than the DTCG-Net. Finally, all the approaches have similar standard deviation ranges on our dataset (DG-CNN ranges from 27\% to 34\%, DCG-Net ranges from 28\% to 30\%, and DTCG-Net ranges from 23\% to 30\%). Results on the augmented datasets UIBK-20 and UIBK-40 (400 images each) tend to confirm that capsule-based methods outperform the CNN-based ones, especially in terms of accuracy.


These results show that the capsule-based networks seem to improve out-of-distribution generalization so that they can be used on unseen object classes. In specific applications, such as the grasping of unknown objects, capsule-based networks could be of great interest since they tend to outperform current state-of-the-art CNN-based networks (Sec.~\ref{chap:introduction}).
    
We believe this is because capsule networks encode better the objects based on their parts and uses the spatial information as well as computing probabilities of an object being present thanks to the nature of the capsule routing \cite{DBLP:journals/corr/abs-1710-09829}. On the other hand, the CNN only tries to locate discriminating features anywhere in the data without taking into account spatial location or part-based object representations.
  
    %



\section{Affordance-based grasping}\label{chap:robot}

To show the effectiveness of our affordance detection network in a real-world scenario, we conducted a grasping experiment with multiple objects and a robot. The goal is to detect the \textit{Grasp} affordance on the observed object and then grasp it. To this end, we used a Franka Emika Panda manipulator mounting an Intel RealSense D415 camera at its end-effector (Fig.~\ref{fig:robot:setup}). As in Sec.~\ref{chap:experiments}, we compare the results obtained with DG-CNN, DCG-Net, and DTCG-Net.

\subsection{Processing pipeline}
\begin{figure}[t!]
    \includegraphics[width=\columnwidth]{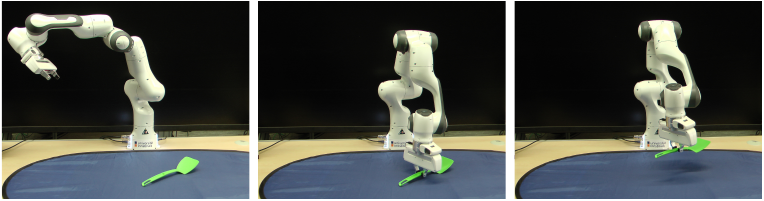}
    \caption{The robot successfully grasps the \texttt{turner}.}
     \label{fig:robot:setup}
\end{figure}

The pipeline used to retrieve a grasping pose from sensory data consists of the following steps. 
%
\begin{enumerate} 
	\item \textit{Capturing the image.} We directly capture an RGB-D image from the Intel Realsense Depth Camera D415.
	\item \textit{Background removal.} 
	The background is removed from the RGB-D image by placing the objects on a blue surface and removing the biggest blue contour within the image using OpenCV~\cite{opencv_library}. We have also tested the plane detection approach proposed by Zhou et al.~\cite{Zhou2018}, but results were not reliable because flat objects were often misinterpreted as being part of the plane. 
	\item \textit{Conversion to point cloud.} We convert the area of interest to a point cloud with the library Open3D~\cite{Zhou2018} and reduce the amount of points to $2048$. 
	\item \textit{Prediction of the network.} We feed the point cloud to the networks in order to get an affordance prediction for the observed object.
	\item \textit{Finding the biggest cluster of type grasp.} To reduce noise and outlier effects, we select the biggest cluster of \textit{Grasp} points. 
	\item \textit{Grasping pose estimation.} Grasping position is computed as the mean of the points in the \textit{Grasp} cluster. For the orientation, we perform a principal component analysis and get 3 axes forming a rotation matrix. 
	\item \textit{Grasping the object.} We use Cartesian impedance control~\cite{Ott2008} to reach the object to grasp. Grasping position is directly fed to the robot. For the orientation, we compute the rotation needed to make the gripper fingers orthogonal to the major axis (longest dimension) of the object. This is obtained by computing the angle between the gripper $x$-axis and the first principal component computed in step 6). With the gripper closed, the robot lifts the object $10\,$cm and we visually determine if it is firmly grasped (Fig.~\ref{fig:robot:setup}). 
\end{enumerate}
%

\subsection{Results}
We used 6 different objects (\texttt{spoon}, \texttt{fork}, \texttt{hammer}, \texttt{knife}, \texttt{cup}, and \texttt{turner}) which we tried to grasp at 4 different orientations. These objects are part of the UIBK dataset, but captured from unseen viewpoints. Each grasping experiment was performed twice to evaluate repeatability. Overall, the robot has to perform 48 grasps for each of the 3 networks. For each run, we checked if the grasp was successful, measured the distance from the ideal grasping point and the deviation (in degrees) for the rotation between the robot and the grasping part. The distance to the grasping point (centroid) is of importance in antipodal grasping because grasping far from the centroid may easily cause the object to tilt. Keeping the correct orientation wrt the grasping part is also important to prevent undesired collisions with the object. We defined a grasp attempt as successful if the object was grasped at a grasping point and the robot was able to pick it up. The grasping point was defined by hand for each object, by taking the geometric center of the biggest grasping part. 

Table~\ref{tab:robot:results} shows the average results over all runs for each network. The success rates denote how many grasp attempts were successful. The distance as well as the angle gives the average distance over all runs for each algorithm shows that DTCG-Net outperforms both DCG-Net and DG-CNN. The robot successfully performed 42 out of 48 grasps with DTCG-Net, 34 out of 48 with DG-CNN, and 32 out of 48 with DCG-Net. In other words, the success rate with DTCG-Net was  $16.7$\% higher when compared to the best baseline approach. Using DTCG-Net, the average distance between actual and ideal grasping points decreased by $13.6$\% ($0.3\,$cm) and the average grasping angle deviation decreased by $46.9$\% ($9.2\,$deg) compared to DG-CNN (the second best network).

The results from this experiment coincide with those shown in Sec.~\ref{chap:experiments} regarding novel objects and viewpoints. We assume that, due to the better affordance detection of the DTCG-Net when confronted with unseen object instances from new viewpoints, the robot is able to perform its grasping task with a higher success rate and with lower error values.

\begin{table}[t]
    \centering
    \caption{Results of the affordance-based grasping experiments.}
    \label{tab:robot:results}
    \resizebox{\columnwidth}{!}{%
    {\renewcommand\arraystretch{1.2} 
 	\begin{tabular}{cccc}
 	    \toprule
         \sc{Network} & \sc{Success Rate} [\%] & \sc{Distance} [cm] & \sc{Angle} [deg] \\\midrule
\sc{DG-CNN} & 70.8               & 2.2                   & 19.6               \\
\sc{DCG-Net} & 66.7               & 2.7                   & 23.3               \\
\sc{DTCG-Net} & 87.5               & 1.9                   & 10.4               \\\bottomrule
    \end{tabular}
    }}
\end{table}

\section{\label{chap:conclusion}Conclusions}
In this work, we have presented the Dynamic-Tree Capsule Graph Network (DTCG-Net) for affordance detection. Different capsule networks were developed for processing 3D point clouds to ensure equivariant representations of capsule vectors w.r.t. rotations, translations, and permutations of inputs  \cite{DCG-Net, quaternion-capsnet}. To the best of our knowledge, we are the first that not only use capsule layers to produce equivariant vectors but also, the proposed DTCG-Net additionally leverages a particular routing algorithm (SDA) to enforce a part-to-whole object inductive bias on the capsules. As demonstrated in the experimental section, this inductive bias improves out-of-distribution generalization capabilities which are very well suited for affordance detection even on unseen object instances.   

We have shown qualitatively that our DTCG-Net capsule network outperforms  DCG-Net in terms of accuracy and mIoU. 
The current state-of-the-art CNN based network DG-CNN still outperforms capsule-based networks at 3D point-cloud classification, including our own DTCG-Net, even if it is by a small margin. However, our DTCG-Net outperforms not only DCG-Net, but also such state-of-the-art CNN when the object instances or views were not presented at training -- which are extremely important in robotic tasks -- due to DTCG-Net's better out-of-distribution generalization capabilities. We additionally evaluated affordance detection with those unseen object instances to show how the aforementioned neural networks can handle unseen objects and new viewpoints and provided a task of affordance detection and grasping. Here, the robot was faced with the grasping of new unseen object instances from new viewpoints. DTCG-Net performed by far the best with the highest success rate of object grasping, the lowest average distance to the ideal grasp point as well as lowest average grasping angle deviation, which shows the superiority of our network in robotic tasks. Behind the success of DTCG-Net over not only DCG-Net, but also the state-of-the-art DTCG-Net at these affordances tasks is the inductive bias implemented through SDA routing. 
This inductive bias allows to detect objects not used during training since the focus is not so much on the highest-performance feature extraction (as in the CNN methods like DG-CNN), but on a meaningful feature extraction for parts representing objects which can be done thanks to Capsule Networks routing algorithms such as SDA, which is specifically designed to build a tree-like structure.

%
Future work would include extending DTCG-Net to provide a direct pose estimation for each detected affordance to avoid the heuristic definition of the grasping orientation. SDA's routing could be also updated to better handle small object parts and evaluate other routing algorithms like recent Variational Bayes \cite{DBLP:journals/corr/abs-1905-11455} or dot-product attention routing \cite{capsules-with-inverted-dot-product-attention-routing}.

\bibliographystyle{IEEEtran}
\bibliography{egbib}

\end{document}